\documentclass[sigconf]{acmart}
\AtBeginDocument{%
  \providecommand\BibTeX{{%
    \normalfont B\kern-0.5em{\scshape i\kern-0.25em b}\kern-0.8em\TeX}}}

\copyrightyear{2021}
\acmYear{2021}
\setcopyright{acmcopyright}
\acmConference[MM '21] {Proceedings of the 29th ACM Int'l Conference on Multimedia}{October 20--24, 2021}{Virtual Event, China.}
\acmBooktitle{Proceedings of the 29th ACM Int'l Conference on Multimedia (MM '21), Oct. 20--24, 2021, Virtual Event, China}
\acmPrice{15.00}
\acmISBN{978-1-4503-8651-7/21/10}
\acmDOI{10.1145/3474085.3475369}

\begin{document}
\fancyhead{}
\title{DSSL: Deep Surroundings-person Separation Learning for Text-based Person Retrieval}


\author{Aichun Zhu}
\authornote{Corresponding author.\vspace{-0.1cm}}
\affiliation{%
  \institution{Nanjing Tech University}
  \city{Nanjing}
  \country{China}}
\email{aichun.zhu@njtech.edu.cn}

\author{Zijie Wang}
\affiliation{%
	\institution{Nanjing Tech University}
	\city{Nanjing}
	\country{China}}
\email{zijiewang9928@gmail.com}

\author{Yifeng Li}
\affiliation{%
	\institution{Nanjing Tech University}
	\city{Nanjing}
	\country{China}}
\email{lyffz4637@163.com}

\author{Xili Wan}
\affiliation{%
	\institution{Nanjing Tech University}
	\city{Nanjing}
	\country{China}}
\email{xiliwan@njtech.edu.cn}

\author{Jing Jin}
\affiliation{%
	\institution{Nanjing Tech University}
	\city{Nanjing}
	\country{China}}
\email{ janeking1015@163.com}

\author{Tian Wang}
\affiliation{%
	\institution{Beihang University}
	\city{Beijing}
	\country{China}}
\email{wangtian@buaa.edu.cn}

\author{Fangqiang Hu}
\affiliation{%
	\institution{Nanjing Tech University}
	\city{Nanjing}
	\country{China}}
\email{ hufq@njtech.edu.cn}

\author{Gang Hua}
\affiliation{%
	\institution{China University of Mining and Technology}
	\city{XuZhou}
	\country{China}}
\email{ghua@cumt.edu.cn}

%
%
%
%
%

\renewcommand{\shortauthors}{Zhu, et al.}


\begin{abstract}
	Many previous methods on text-based person retrieval tasks are devoted to learning a latent common space mapping, with the purpose of extracting modality-invariant features from both visual and textual modality. Nevertheless, due to the complexity of high-dimensional data, the unconstrained mapping paradigms are not able to properly catch discriminative clues about the corresponding person while drop the misaligned information. Intuitively, the information contained in visual data can be divided into person information (PI) and surroundings information (SI), which are mutually exclusive from each other. To this end, we propose a novel Deep Surroundings-person Separation Learning (DSSL) model in this paper to effectively extract and match person information, and hence achieve a superior retrieval accuracy. A surroundings-person separation and fusion mechanism plays the key role to realize an accurate and effective surroundings-person separation under a mutually exclusion constraint. In order to adequately utilize multi-modal and multi-granular information for a higher retrieval accuracy, five diverse alignment paradigms are adopted. Extensive experiments are carried out to evaluate the proposed DSSL on CUHK-PEDES, which is currently the only accessible dataset for text-base person retrieval task. DSSL achieves the state-of-the-art performance on CUHK-PEDES. To properly evaluate our proposed DSSL in the real scenarios, a Real Scenarios Text-based Person Reidentification (RSTPReid) dataset is constructed to benefit future research on text-based person retrieval, which will be publicly available at \url{https://github.com/NjtechCVLab/RSTPReid-Dataset}.
\end{abstract}

\ccsdesc[500]{Information systems~Image search}
\ccsdesc[500]{Computing methodologies~Object identification}

\keywords{person retrieval, text-based person re-identification, cross-modal retrieval, surroundings-person separation}

\maketitle

\section{Introduction}

Person retrieval is a basic task in the field of video surveillance, which aims to identify the corresponding pedestrian in a large-scale person image database with a given query. Current researches of person retrieval chiefly focus on image-based person retrieval \cite{yi2014deepreid, IAM2019CVPR, SecondOrder2019} (aka. person re-identification), which may sometimes suffer from lacking query images of the target pedestrian in practical application. Considering that in most of the real-world scenes, textual description queries are much more accessible, text-based person retrieval \cite{Shuang2017Person, li2017identity, niu2020improving, Jing2018Pose, ARL, wang2020img, mm2019graphreid} has drawn remarkable attention for its effectiveness and applicability.

\begin{figure}[h]
	\centering
	\includegraphics[width=0.9\linewidth]{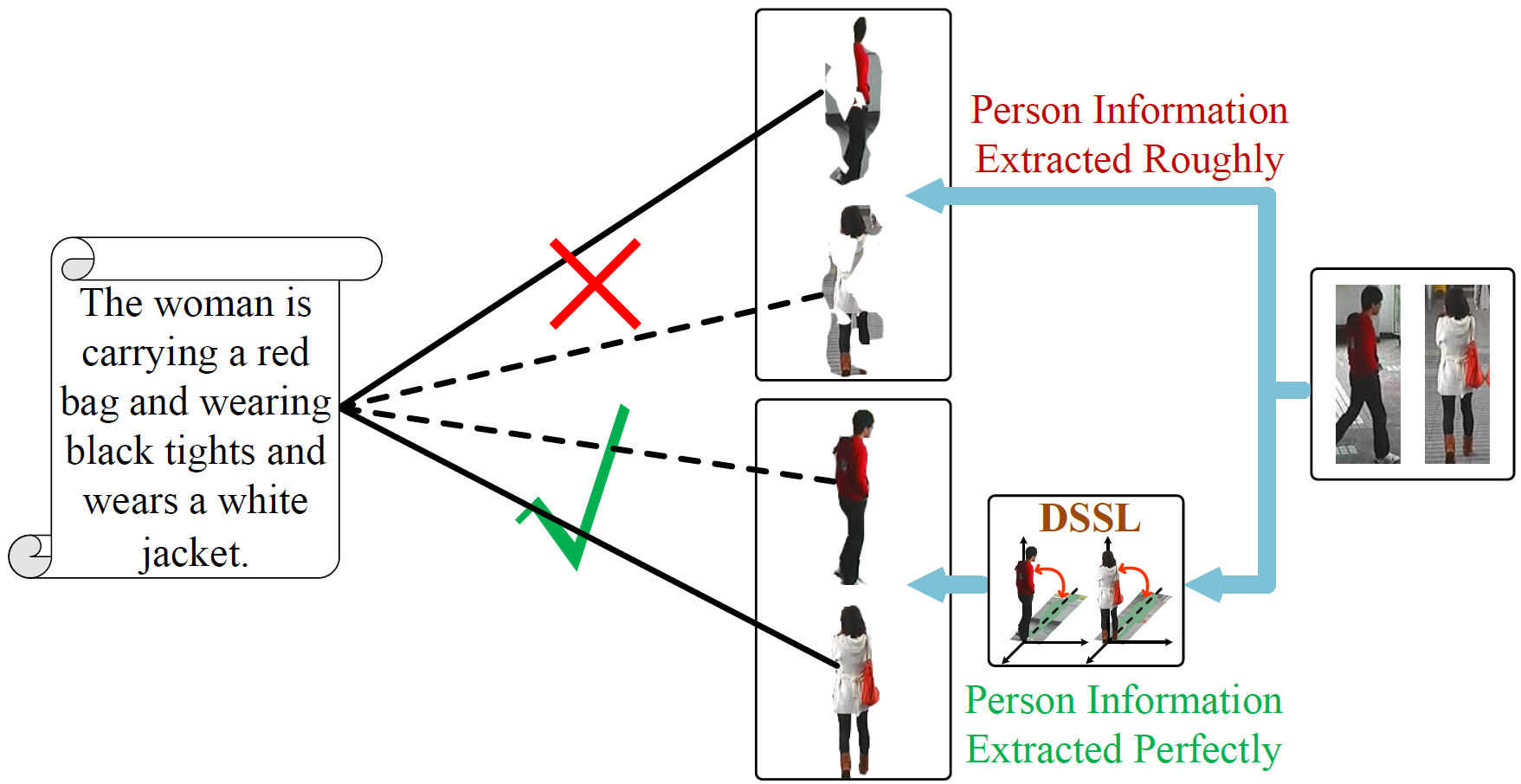}
	\caption{Due to the complexity of high-dimensional data, roughly extracting the person information without proper constraints may miss key clues while fail to drop redundant inferences, which further leads to a mismatched case. Within our proposed Deep Surroundings-person Separation Learning (DSSL) model, the person information is properly separated with the surroundings information, which hence gives a superior retrieval performance.}
	\label{fig:motivation}
\end{figure}

\begin{figure*}[h]
	\centering
	\includegraphics[width=0.75\linewidth]{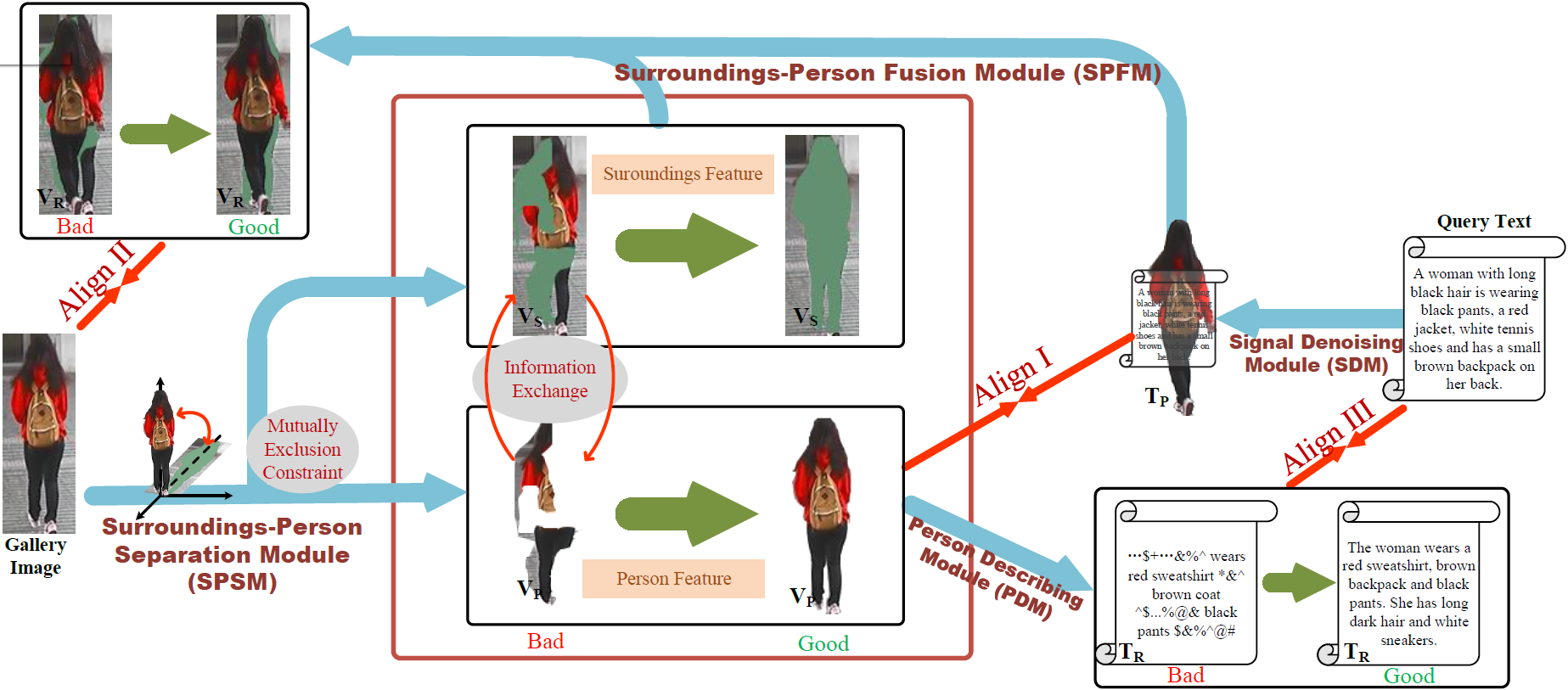}
	\caption{Illustration of the complementary relationship among $Align I$, $Align II$ and $Align III$ when training DSSL. During the training process, $Align I$ and $Align III$ will form a constraint which forces more complete information about the person to be contained in the person feature $V_{P}$. Based on the mutually exclusion precondition, person information in the surroundings feature $V_{S}$ will accordingly be taken away into $V_{P}$. Meanwhile, $Align II$ is conducted by putting the described person into the surroundings of the gallery person, which requires the surroundings information to be properly included in $V_{S}$ while peeled off in $V_{P}$. As a result, these three alignments work in complementary to guide the correct information exchange under a mutually exclusion constraint, which finally leads to the change of extracted information from bad to good and an accurate and effective surroundings-person separation.}
	\Description{Illustration of the complementary relationship among $Align I$, $Align II$ and $Align III.$}
	\label{fig:training}
\end{figure*}

As text-based person retrieval involves processing multi-modal data, it can be deemed as a specific subtask of cross-modal retrieval \cite{yan2015DCCA, karpathy2015DVSA, nam2017DAN, liu2017RRFNet, sun2019supervisedhashingsigir, SCAN}. Nevertheless, instead of containing various categories of objects in an image, each image cared by text-based person retrieval contains just one certain pedestrian. The textual description queries, meanwhile, offer much more details about the corresponding person rather than roughly mention the objects in an image. Owing to the particularity of text-based person retrieval, many previous methods proposed on general cross-modal retrieval benchmarks (e.g. Flickr30K \cite{plummer2015flickr30k} and MSCOCO \cite{MSCOCO}) generalize on it poorly. In addition, CUHK-PEDES \cite{Shuang2017Person} is currently the only accessible dataset for text-base person retrieval. It is large in scale and contains images collected from various re-identification datasets under different scenes, view points and camera specifications. Nevertheless, images of each specific person are mostly caught by a same camera under similar conditions of time and space, which is not consistent with the real application scenarios. Therefore, we construct a Real Scenarios Text-based Person Reidentification (RSTPReid) dataset based on MSMT17 \cite{MSMT17} to further train and evaluate the performance of our work, which also benefit future research. For each person, RSTPReid pools 5 images caught by 15 different cameras with complex both indoor and outdoor scene transformations and backgrounds in various periods of time, which makes RSTPReid much more challenging and more adaptable to real scenarios. Extensive experiments on RSTPReid and CUHK-PEDES can better validate the promising accuracy and efficiency of our work.

The major challenge of text-based person retrieval is to effectively extract and match features from both raw images and textual descriptions. Many previous methods \cite{ARL, mm2019graphreid, niu2020improving, Jing2018Pose, wang2020img} are devoted to learning a latent common space mapping, with the purpose of extracting modality-invariant features from both the visual and textual modalities. These proposed approaches are mainly based on the assumption that through a latent common space mapping, the intersection of information carried by the two modalities, namely, information of the targeting person can be retained into extracted modality-invariant common features. Nevertheless, due to the complexity of high-dimensional data, the unconstrained mapping paradigms are not able to properly catch discriminative clues about the corresponding person while drop the misaligned information (shown in Fig. \ref{fig:motivation}).

Intuitively, information contained in visual data can be divided into person information (PI) and surroundings information (SI), which are \textbf{mutually exclusive} from each other. Meanwhile, the given textual description queries commonly describe the gender, appearance, clothing, carry-on items, possible movement, etc. of a certain pedestrian. In most of the real scenarios, the describer who offers a query nearly knows nothing about what kind of surroundings the target person is exactly in when captured by surveillance cameras, where the light conditions, viewpoints, etc. can be varied. Therefore, the given textual description basically contains only person information and there is no surroundings information included. On account of the structure of natural language sentences, noise signals (NS) like semantically irrelevant words and incorrect grammar are also inevitably included. Based on the above discussion, an efficient algorithm to accurately separate person and surroundings information in visual data and properly denoise features extracted from textual data is essential to enhance the retrieval performance.

\begin{figure*}[h]
	\centering
	\includegraphics[width=0.9\linewidth]{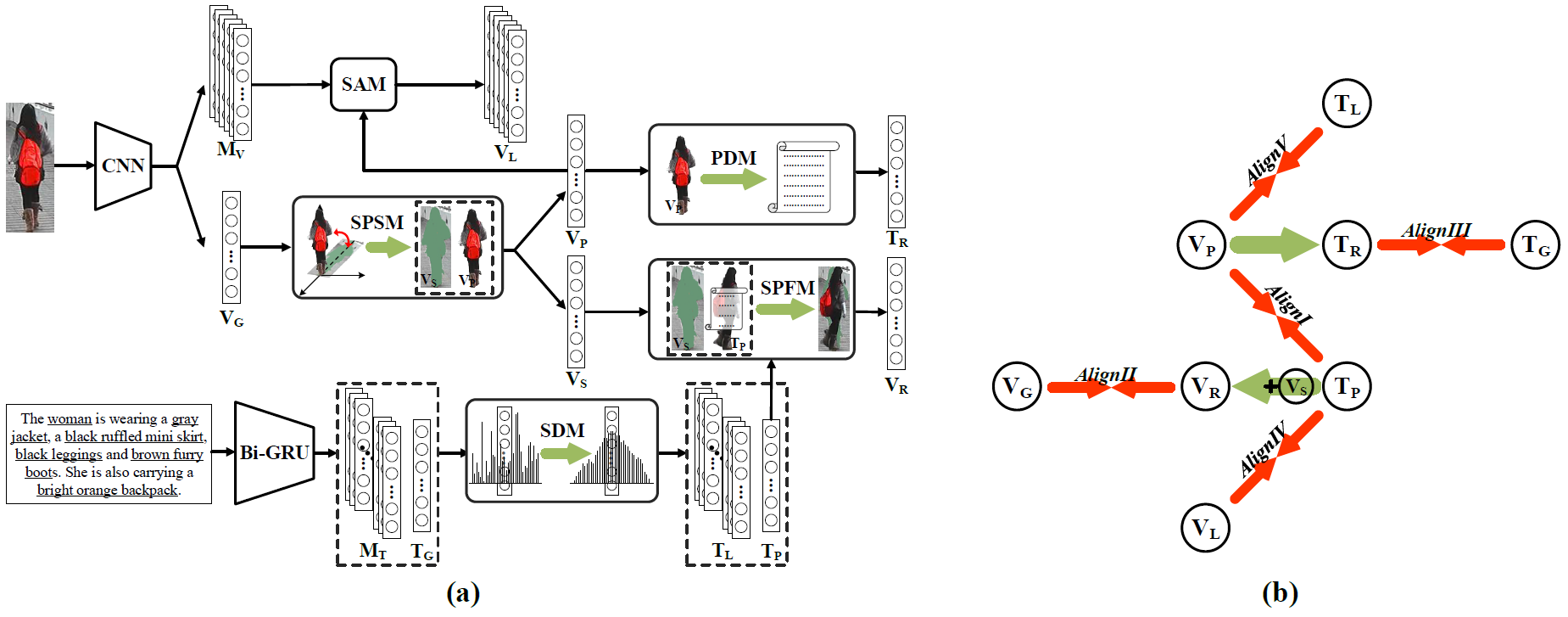}
	\caption{The overall framework of the proposed Deep Surroundings-person Separation Learning (DSSL) model. A surroundings-person separation and fusion mechanism plays the key role to realize an accurate and effective surroundings-person separation under a mutually exclusion constraint. (a) Illustration of the proposed feature extraction procedure in DSSL. (b) Illustration of the five diverse alignment paradigms adopted to adequately utilize multi-modal and multi-granular information and hence improve the retrieval accuracy. $V_{G}$/$T_{G}$, $V_{P}$/$T_{P}$, $V_{R}$/$T_{R}$ and $V_{L}$/$T_{L}$ denote the extracted visual/textual global, person, reconstructed and local features, respectively.}
	\Description{Deep Surroundings-person Separation Learning (DSSL) model.}
	\label{fig:model}
\end{figure*}

To this end, we propose a novel Deep Surroundings-person Separation Learning (DSSL) model in this paper to effectively extract and match person information, and hence achieve a superior retrieval accuracy. DSSL takes raw images and textual descriptions as input and first extracts global and fine-grained local information from both modalities. As shown in Fig. \ref{fig:training}, DSSL aims to properly separate surroundings and person information. To achieve this goal, a novel Surroundings-Person Separation Module (SPSM) is proposed to split the visual information as person and surroundings features (denoted as $V_{P}$ and $V_{S}$) in a mutually exclusive manner. Then we adopt a Signal Denoising Module (SDM) to denoise and refine the extracted person feature (denoted as $T_{P}$) from the textual modality. As discussed above, ideally the person features $V_{P}$ and $T_{P}$ are purely about the target person without modality-specific interference. Hence the alignment between them ($Align I$) can be regarded as matching the pedestrian cut out of the gallery image with the pedestrian in mind of the describer. In addition, through a proposed Surroundings-Person Fusion Module (SPFM), $T_{P}$ is fused with $V_{S}$ and reconstructed into the visual modality as $V_{R}$. Then an alignment between $V_{R}$ and the visual feature before partitioned by SPSM ($Align II$) is conducted, which can be viewed as placing the described person into the same surroundings as the gallery person and then matching it with the complete gallery image including the surroundings in the visual modality space. Besides, a Person Describing Module (PDM) is employed to reconstruct $V_{P}$ into textual modality as $T_{R}$, which is then aligned with the non-refined textual feature ($Align III$). This proposed alignment can be regarded as describing the person in the gallery image with a text and then matching the text with the given query sentence in the textual modality space. Due to the mutually exclusion constraint in SPSM, $V_{P}$ and $V_{S}$ are orthogonal to each other, so the visual information is distributed between them without overlap. As shown in Fig. \ref{fig:training}, during the training process, $Align I$ and $Align III$ will form a constraint which forces $V_{P}$ to contain more complete information about the person. Based on the mutually exclusion precondition, person information in $V_{S}$ will accordingly be taken away into $V_{P}$. Meanwhile, $Align II$ is conducted by putting the described person into the surroundings of the gallery person, which requires the surroundings information to be properly included in $V_{S}$ while peeled off in $V_{P}$. As a result, these three alignments work in complementary to guide the correct information exchange between $V_{P}$ and $V_{S}$ under a mutually exclusion constraint, and finally lead to an accurate and effective surroundings-person separation. To adequately exploit fine-grained clues, a cross-modal attention (CA) mechanism \cite{niu2020improving, wang2020img} is utilized to further align a local feature matrix extracted from one modality with the global feature in the other ($Align IV$ and $Align V$ shown in Fig. \ref{fig:model}).

Our contributions can be summarized as five folds: (1) A novel Deep Surroundings-person Separation Learning (DSSL) model is proposed to properly extract and match person information. A proposed surroundings-person separation and fusion mechanism plays the key role to realize an accurate and effective surroundings-person separation under a mutually exclusion constraint. (2) Five diverse alignment paradigms are adopted to adequately utilize multi-modal and multi-granular information and hence improve the retrieval accuracy. (3) A Signal Denoising Module (SDM) is employed to denoise and refine the extracted person feature from the textual modality. (4) Extensive experiments are carried out to evaluate the proposed DSSL on CUHK-PEDES \cite{Shuang2017Person}. DSSL outperforms previous methods and achieves the state-of-the-art performance on CUHK-PEDES. (5) A Real Scenarios Text-based Person Reidentification (RSTPReid) dataset is constructed to benefit future research on text-based person retrieval, which will be publicly available.

\section{Related Works}
\subsection{Person Re-identification}
Person re-identification has drawn increasing attention in both academical and industrial fields, and deep learning methods generally plays a major role in current state-of-the-art works. Yi et al. \cite{yi2014deepreid} firstly proposed deep learning methods to match people with the same identification. Hou et al. \cite{IAM2019CVPR} proposed an Interaction-and-Aggregation (IA) Block, which consists of Spatial Interaction-and-Aggregation (SIA) and Channel Interaction-and-Aggregation (CIA) Modules to strengthen the representation capability of the deep neural network. Xia et al. \cite{SecondOrder2019} proposed the Second-order Non-local Attention (SONA) Module to learn local/non-local information in a more end-to-end way.

\subsection{Text-based Person Retrieval}
Text-based person retrieval aims to search for the corresponding pedestrian image according to a given text query. This task is first put forward by Li et al. \cite{Shuang2017Person} and they take an LSTM to handle the input image and text. An efficient patch-word matching model \cite{Chen2018} is proposed to capture the local similarity between image and text. Jing et al. \cite{Jing2018Pose} utilize pose information as soft attention to localize the discriminative regions. Niu et al. \cite{niu2020improving} propose a Multi-granularity Image-text Alignments (MIA) model exploit the combination of multiple granularities. Nikolaos et al. \cite{ARL} propose a Text-Image Modality Adversarial Matching approach (TIMAM) to learn modality-invariant feature representation by means of adversarial and cross-modal matching objectives. Besides that, in order to better extract word embeddings, they employ the pre-trained publicly-available language model BERT. An IMG-Net model is proposed by Wang et al. \cite{wang2020img} to incorporate inner-modal self-attention and cross-modal hard-region attention with the fine-grained model for extracting the multi-granular semantic information. Liu et al. \cite{mm2019graphreid} generate fine-grained structured representations from images and texts of pedestrians with an A-GANet model to exploit semantic scene graphs. A new approach CMAAM is introduced by Aggarwal et al. \cite{aggarwal2020text} which learns an attribute-driven space along with a class-information driven space by introducing extra attribute annotation and prediction. Zheng et al. \cite{zheng2020hierarchical} propose a Gumbel attention module to alleviate the matching redundancy problem and a hierarchical adaptive matching model is employed to learn subtle feature representations from three different granularities. Recently, the NAFS proposed by Gao et al. \cite{gao2021contextual} is designed to extract full-scale image and textual representations with a novel staircase CNN network and a local constrained BERT model. Besides, a multi-modal re-ranking algorithm by comparing the visual neighbors of the query to the gallery (RVN) is utilized to further improve the retrieval performance.

\section{Methodology}
In this section, we describe the proposed Deep Surroundings-person Separation Learning (DSSL) model in detail (shown in Fig. \ref{fig:model}), which consists of a Surroundings-Person Separation Module (SPSM), a Surroundings-Person Fusion Module (SPFM), a Signal Denoising Module (SDM), a Person Describing Module (PDM) and a Salient Attention Module (SAM).

\subsection{Feature Extraction And Refinement}
\label{sec:feature_extraction}
\subsubsection{Feature Extraction}
We utilize a ResNet-50 \cite{ResNet} backbone pretrained on ImageNet to extract global/local visual features from a given image $I$. To obtain the global feature $V_{G} \in \mathbb{R}^{p}$, the feature map before the last pooling layer of ResNet-50 is down-scaled to a vector $\in \mathbb{R}^{1 \times 1 \times 2048}$ with an average pooling layer and then passed through a group normalization (GN) layer followed by a fully-connected (FC) layer. In the local branch, the same feature map is first horizontally $k$-partitioned by pooling it to $k \times 1 \times 2048$, and then the local strips are separately passed through a GN and two FCs with a ReLU layer between them to form $k$ $p$-dim vectors, which are finally concatenated to obtained the local visual feature matrix $M_{V} \in \mathbb{R}^{k \times p}$.

For textual feature extraction, we take a whole sentence and the $n$ phrases extracted from it as textual materials, which are handled by a bi-directional GRU (bi-GRU). The last hidden states of the forward and backward GRUs are concatenated to give global/local $p$-dim feature vectors. The $p$-dim vector got from the whole sentence is passed through a GN followed by an FC to form the global textual feature $T_{G} \in \mathbb{R}^{p}$. With each certain input phrase, the corresponding output $p$-dim vector is processed consecutively by a GN and two FCs with a ReLU layer between them and then concatenated with each other to form the local textual feature matrix $M_{T} \in \mathbb{R}^{n \times p}$.

\subsubsection{Textual Person Information Refinement}
To further remove the noise signals in the textual data, so as to refine the extracted person information, the global feature vector $T_{G}$ and local feature vectors in $M_{T}$ are separately handled by a \textbf{Signal Denoising Module (SDM)}. With a fixed zeroing ratio $r$, a fixed number of elements in an input vector is set to zero \cite{vincent2008denoisingAE}. And then the processed vector is reconstructed following an autoencoder manner to obtain the textual person feature vector $T_{P}$ and the local textual person feature matrix $T_{L}$:
\begin{equation}
	y = Dec(Enc(Z(x, r))),
\end{equation}
where $Z(x, r)$ denotes the zero setting operation with ratio $r$, $x \in \{T_{G}\} \cup M_{T}$ and $y \in \{T_{P}\} \cup T_{L}$. With the zeroing and reconstruction mechanism, the input vectors are required to fully retain effective information while discarding redundant noise signals. The reconstruction loss of SDM is defined as:
\begin{equation}
	L_{SDM} = L_{rank}(T_G, T_P) + \sum_{i=1}^{k} L_{rank}((M_{T})_{i}, (T_{L})_{i}),
\end{equation}
where $(M_{T})_{i}$ and $(T_{L})_{i}$ denote the $i$-th vector in matrices $M_{T}$ and $T_{L}$. Rather than being superficially look-alike, the denoised vector ought to be properly matched with the original vector because of the special nature of a retrieval task. Therefore, instead of utilizing the traditional Euclidean Distance to guide the reconstruction, a triplet ranking loss is adopted:
\begin{multline}
	L_{rank}(x_1, x_2) = \sum_{\widehat{x_2}} max\{\alpha - S(x_1, x_2) + S(x_1, \widehat{x_2}), 0\}\\
	+ \sum_{\widehat{x_1}} max\{\alpha - S(x_1, x_2) + S(\widehat{x_1}, x_2), 0\},
\end{multline}
to more accurately constrain the matched pairs to be closer than the mismatched pairs with a margin $\alpha$, where $(x_1, \widehat{x_2})$ or $(\widehat{x_1}, x_2)$ denotes a mismatched pair and $S(\cdot, \cdot)$ is the cosine similarity between two vectors. Instead of using the furthest positive and closest negative sampled pairs, we adopt the sum of all pairs within each mini-batch when computing the loss following \cite{faghri2017vse++}.

\subsection{Deep Surroundings-Person Separation Learning}
\label{sec:feature_alignment}
As shown in Fig. \ref{fig:model} (b), five alignment paradigms are adopted to adequately utilize multi-modal and multi-granular information for a robust Deep Surroundings-Person Separation Learning process, thereby improving the retrieval accuracy.
\subsubsection{Align I}
To process the visual data, the person feature $V_{P}$ and surroundings feature $V_{S}$ are separated through a \textbf{Surroundings-Person Separation Module (SPSM)}, which is implemented as two paralleled multi-layer perceptrons (MLP) (with the feature dimension conversion as $p \rightarrow 2p \rightarrow p$) followed by a $tanh$ layer:
\begin{equation}
	V_{P}, V_{S} = SPSM(V_{G}).
\end{equation}
The person features extracted from both modalities are first aligned. The alignment loss for $Align I$ is
\begin{equation}
	L_{AlignI} = L_{rank}(V_{P}, T_{P}).
\end{equation}
Besides, a \textbf{Mutually Exclusion Constraint (MEC)} is proposed to ensure that $V_{P}$ and $V_{S}$ are orthogonal to each other and the visual information is distributed between them without overlap. Let $P = \{V^{i}_{P}\}_{i = 1}^{B} \in \mathbb{R}^{B \times p}$ and $S = \{V^{i}_{S}\}_{i = 1}^{B} \in \mathbb{R}^{B \times p}$ denote matrices whose rows are person and surroundings features in a training batch, respectively, where $B$ is the batch size, and then the mutually exclusion loss is
\begin{equation}
	L_{MEC} = \Vert P^{T}S \Vert.
\end{equation}

\subsubsection{Align II}
With a proposed \textbf{Surroundings-Person Fusion Module (SPFM)}, $T_{P}$ is fused with $V_{S}$ and reconstructed into the visual modality as $V_{R}$:
\begin{equation}
	V_{R} = SPFM(T_{P}, V_{S}),
\end{equation}
which is then aligned with $V_{G}$ and the alignment loss for $AlignII$ is
\begin{equation}
	L_{AlignII} = L_{rank}(V_{G}, V_{R}).
\end{equation}
$SPFM$ first combine the two input vectors by addition or concatenation (compared in Section \ref{sec:exp_SPSM+SPFM}), and then the combined feature is processed by an MLP similar to $SPSM$.

\subsubsection{Align III}
A \textbf{Person Describing Module (PDM)}, which is implemented as an MLP with a $tanh$ activation function is employed to reconstruct $V_{P}$ into the textual modality as $T_{R}$ and then aligned with $T_{G}$:
\begin{equation}
	T_{R} = PDM(V_{P}).
\end{equation}
The alignment loss for $AlignIII$ is
\begin{equation}
	L_{AlignIII} = L_{rank}(T_{G}, T_{R}).
\end{equation}

\subsubsection{Align IV}
A \textbf{Salient Attention Module (SAM)} is first employed to highlight person information in the local visual feature matrix $M_{V}$:
\begin{equation}
	(V_{L})_{i} = Sigmoid(W_{2}(GN(ReLU(W_{1}(V_{P}) + b_{1}))) + b_{2}) \cdot (M_{V})_{i},
\end{equation}
where $GN$ denotes the group normalization layer while $W_{1}, W_{2}$ and $b_{1} , b_{2}$ denote the linear transformation. To adequately exploit fine-grained clues, a cross-modal attention (CA) mechanism \cite{niu2020improving, wang2020img} is utilized to align $V_{L}$ with the textual person feature $T_{P}$ and form a $p$-dim vector:
\begin{equation}
	CA(V_{L}, T_{P}) = \sum\limits_{\alpha^{i}_{V} > \frac{1}{k}} \alpha^{i}_{V} (V_{L})_{i},
	\alpha^{i}_{V} = \frac{exp(cos((V_{L})_{i}, T_{P}))}{\sum_{j=1}^{k} exp(cos((V_{L})_{j}, T_{P}))},
\end{equation}
where $\alpha^{i}_{V}$ represents the relation between the $i$-th local visual part and textual person feature. And the alignment loss for $Align IV$ is
\begin{equation}
	L_{AlignIV} = L_{rank}(CA(T_{P}, V_{L}), T_{P}).
\end{equation}

\subsubsection{Align V}
Similar with $Align IV$, the alignment loss for $Align V$ is
\begin{equation}
	L_{AlignV} = L_{rank}(CA(V_{P}, T_{L}), V_{P}),
\end{equation}
\begin{equation}
	CA(T_{L}, V_{P}) = \sum\limits_{\alpha^{i}_{T} > \frac{1}{n}} \alpha^{i}_{T} (T_{L})_{i},
	\alpha^{i}_{T} = \frac{exp(cos((T_{L})_{i}, V_{P}))}{\sum_{j=1}^{n} exp(cos((T_{L})_{j}, V_{P}))}.
\end{equation}

\subsection{Loss Function for Training}
\label{sec:loss}
The complete training process includes 2 stages.

\subsubsection{Stage-1}
We first fix the parameters of the ResNet-50 backbone and train the left feature extraction part of DSSL with the identification (ID) loss 
\begin{equation}
	L_{id}(X) =  -log(softmax(W_{id} \times GN(X))
\end{equation}
to cluster person images into groups according to their identification, where $W_{id} \in \mathbb{R}^{Q \times p}$ is a shared transformation matrix implemented as a FC layer without bias and $Q$ is the number of different people in the training set. As global features can provide more complete information for clustering, only $V_{G}$ and $T_{G}$ are utilized here: 
\begin{equation}
	L_{ID1} = L_{id}(V_{G}) + L_{id}(T_{G}).
\end{equation}
And the entire loss in Stage-1 is
\begin{equation}
	L_{Stage1} = L_{ID1}.
\end{equation}

\subsubsection{Stage-2}
In this stage, all the parameters of DSSL are fine-tuned together. Here the ID loss is also employed to ensure that the person features and reconstructed features can be correctly related to the corresponding person:
\begin{equation}
	L_{ID2} = L_{ID1} + L_{id}(V_{P}) + L_{id}(T_{P}) + L_{id}(V_{R}) + L_{id}(T_{R}).
\end{equation}
The five alignment losses are utilized to improve retrieval accuracy:
\begin{equation}
	L_{Alignment} = L_{AlignI} + L_{AlignII} + L_{AlignIII} + L_{AlignIV} + L_{AlignV}.
\end{equation}
Along with the mutually exclusion loss, the entire loss in Stage-2 is
\begin{equation}
	L_{Stage2} = L_{ID2} + L_{Alignment} + L_{MEC}.
\end{equation}

\section{Experiments}
\subsection{Experimental setup}
\subsubsection{Dataset and metrics}
Our approach is evaluated on two challenging datasets: CUHK-PEDES \cite{Shuang2017Person} and our proposed Real Scenario Text-based Person Re-identification (RSTPReid) dataset.

\paragraph{CUHK-PEDES} Previously, CUHK-PEDES \cite{Shuang2017Person} is the only available dataset for text-based person retrieval task. Following the official data split approach, the training set contains 34054 images, 11003 persons and 68126 textual descriptions. The validation set contains 3078 images, 1000 persons and 6158 textual descriptions while the testing set has 3074 images, 1000 persons and 6156 descriptions. Every image generally has two descriptions, and each sentence is commonly no shorter than 23 words. After dropping words that appear less than twice, the word number is 4984.

\begin{figure}[h]
	\centering
	\includegraphics[width=\linewidth]{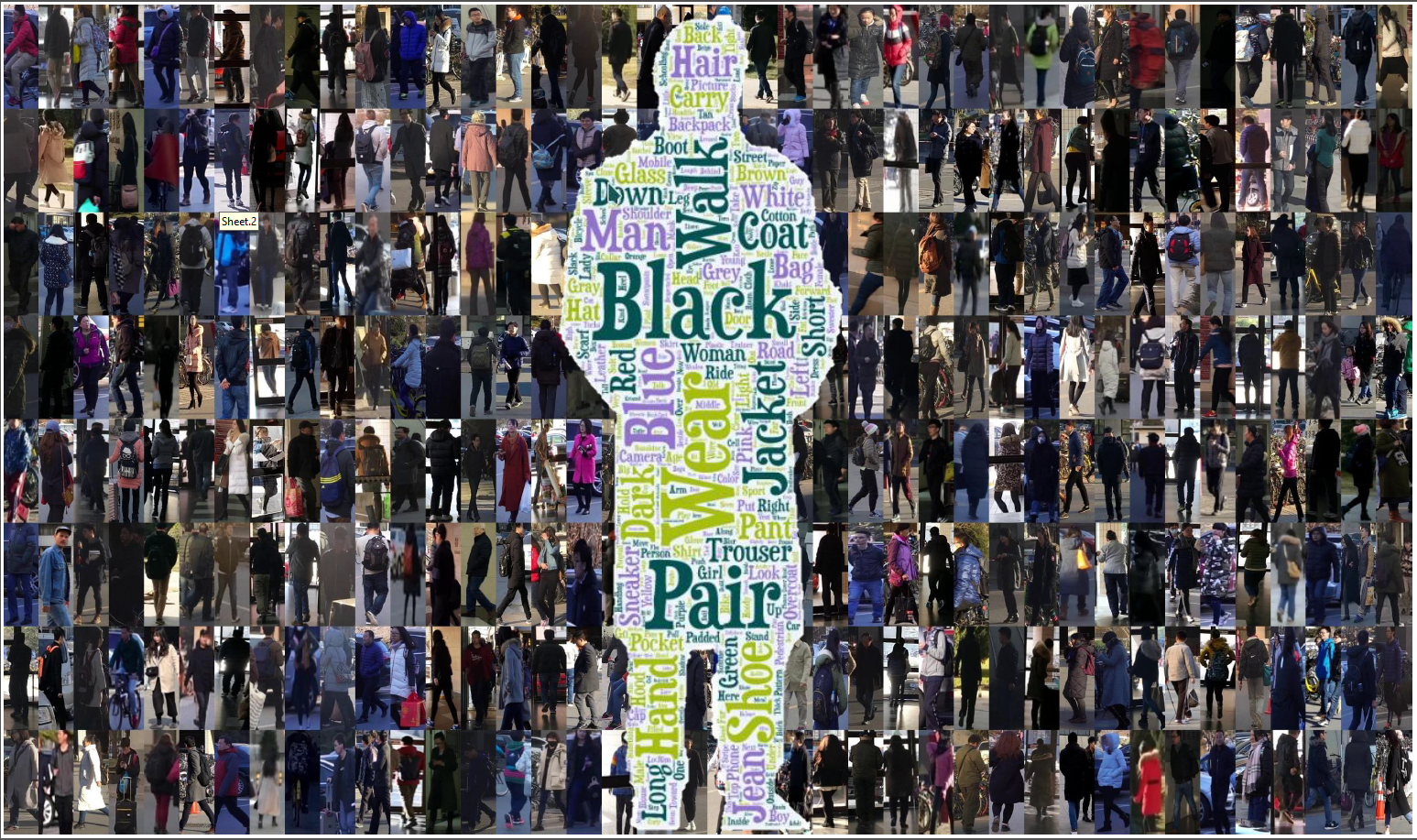}
	\caption{High-frequency words and person images in our constructed RSTPReid dataset.}
	\Description{High-frequency words and person images in RSTPReid.}
	\label{fig:rstpreid}
\end{figure}

\paragraph{RSTPReid} To properly handle real scenarios, we construct a new dataset called Real Scenario Text-based Person Re-identification (RSTPReid) based on MSMT17 \cite{MSMT17}. RSTPReid contains 20505 images of 4,101 persons from 15 cameras. Each person has 5 corresponding images taken by different cameras and each image is annotated with 2 textual descriptions. For data division, 3701, 200 and 200 identities are utilized for training, validation and testing, respectively. Each sentence is no shorter than 23 words. After dropping words that appear less than twice, the word number is 2204. High-frequency words and examples of person images in RSTPReid are shown in Fig. \ref{fig:rstpreid}.

The performance is evaluated by the top-$k$ accuracy. Given a query description, all test images are ranked by their similarities with this sentence. If any image of the corresponding person is contained in the top-$k$ images, we call this a successful search. The top-1, top-5, and top-10 accuracy for all experiments are reported.

\subsubsection{Implementation details}
The feature dimension $p$ is set to $1024$ and the number of local strips $k$ is set to 6. The total number of phrases $n$ obtained from each sentence is kept flexible with an upper bound $26$, which are obtained with the Natural Language ToolKit (NLTK) by syntactic analysis, word segmentation and part-of-speech tagging. We adopt an Adam optimizer to train DSSL with a batch size of 32. The margin $\alpha$ of ranking losses is set to $0.2$. In training stage-1, DSSL is trained with a learning rate of $1\times10^{-3}$ for $10$ epochs with the ResNet-50 backbone fixed. In stage-2, the learning rate is initialized as $2\times10^{-4}$ to optimize all parameters including the visual backbone for extra $30$ epochs. The learning rate is down-scaled by $\frac{1}{10}$ every $10$ epochs. $\lambda_{1}$ and $\lambda_{2}$ in $Sim_{TI}$ are both set to $0.5$. In testing and real application, a cross-modal re-ranking scheme (RR) based on \cite{wang2019matchingRR} is employed to further improve the retrieval accuracy in testing and real application.

\begin{table*}
	\caption{Ablation analysis of the five alignment paradigms in DSSL on CUHK-PEDES and RSTPReid.}
	\label{tab:alignment}
	\begin{tabular}{c |ccccc|ccc | ccc}
		\toprule
		- & - & - & - & - & - & & CUHK-PEDES & & & RSTPReid &\\
		\midrule
		Baseline & $Align I$ & $Align II$ & $Align III$ & $Align IV$ & $Align V$ & Top-1 & Top-5 & Top-10 & Top-1 & Top-5 & Top-10 \\
		\midrule		
		$\checkmark$ & $\times$ & $\times$ & $\times$ & $\times$ & $\times$ & 52.42 & 76.06 & 84.94 & 26.31 & 46.90 & 58.33 \\
		\midrule
		$\times$ & $\checkmark$ & $\times$ & $\times$ & $\times$ & $\times$ & 56.18 & 79.56 & 86.45 & 28.74 & 50.88 & 61.69 \\
		$\times$ & $\times$ & $\checkmark$ & $\times$ & $\times$ & $\times$ & 55.01 & 77.68 & 85.25 & 26.83 & 49.55 & 59.91 \\
		$\times$ & $\times$ & $\times$ & $\checkmark$ & $\times$ & $\times$ & 54.56 & 78.49 & 85.64 & 27.01 & 50.02 & 60.67 \\
		$\times$ & $\times$ & $\times$ & $\times$ & $\checkmark$ & $\times$ & 54.65 & 78.30 & 85.51 & 26.73 & 50.71 & 60.25 \\
		$\times$ & $\times$ & $\times$ & $\times$ & $\times$ & $\checkmark$ & 51.01 & 75.47 & 83.01 & 25.73 & 48.99 & 59.82 \\
		$\times$ & $\checkmark$ & $\checkmark$ & $\times$ & $\times$ & $\times$ & 57.31 & 79.56 & 86.42 & 29.23 & 51.55 & 61.77 \\
		$\times$ & $\checkmark$ & $\times$ & $\checkmark$ & $\times$ & $\times$ & 56.73 & 79.21 & 86.65 & 28.87 & 51.81 & 62.43 \\
		$\times$ & $\times$ & $\checkmark$ & $\checkmark$ & $\times$ & $\times$ & 57.08 & 79.11 & 86.06 & 29.51 & 51.89 & 62.22\\
		$\times$ & $\checkmark$ & $\checkmark$ & $\checkmark$ & $\times$ & $\times$ & 58.86 & 79.70 & 86.95 & 31.00 & 53.83 & 62.63 \\
		$\times$ & $\checkmark$ & $\times$ & $\times$ & $\checkmark$ & $\checkmark$ & 58.19 & 79.41 & 86.52 & 30.81 & 53.67 & 62.71 \\
		\midrule
		$\times$ & $\checkmark$ & $\checkmark$ & $\checkmark$ & $\checkmark$ & $\checkmark$ & \textbf{59.98} & \textbf{80.41} & \textbf{87.56} & \textbf{32.43} & \textbf{55.08} & \textbf{63.19} \\
		\bottomrule
	\end{tabular}
\end{table*}

\begin{table}
	\caption{Ablation analysis of the mutually exclusion constraint (MEC), surroundings-person separation and fusion (SPSM + SPFM), salient attention module (SAM) and signal denoising module (SDM) on CUHK-PEDES.}
	\label{tab:sp}
	\begin{tabular}{cc|c|c|ccc}
		\toprule
		$MEC$ & $SPSM + SPFM$ & $SAM$ & $SDM$ & Top-1 & Top-5 & Top-10 \\
		\midrule
		$\times$ & $\times$ & $\checkmark$ & $\checkmark$ & 55.52 & 77.62 & 85.19 \\
		$\times$ & $\checkmark$ & $\checkmark$ & $\checkmark$ & 57.31 & 79.33 & 86.86 \\
		\midrule
		$\checkmark$ & $\checkmark$ & $\checkmark$ & $\checkmark$ & \textbf{59.98} & \textbf{80.41} & \textbf{87.56} \\
		\midrule
		$\checkmark$ & $\checkmark$ & $\times$ & $\checkmark$ & 57.95 & 79.89 & 87.20 \\
		$\checkmark$ & $\checkmark$ & $\checkmark$ & $\times$ & 57.46 & 79.67 & 87.19 \\
		\bottomrule
	\end{tabular}
\end{table}

\begin{table}
	\caption{Performance comparison of the feature combination method utilized in SPFM on CUHK-PEDES.}
	\label{tab:catadd}
	\begin{tabular}{c|ccc}
		\toprule
		Method & Top-1 & Top-5 & Top-10 \\
		\midrule
		Addition & \textbf{59.98} & 80.41 & \textbf{87.56} \\
		Concatenation & 59.54 & \textbf{80.45} & 87.17 \\
		\bottomrule
	\end{tabular}
\end{table}

\begin{table}
	\caption{Performance comparison of zeroing rate $r$ in SDM on CUHK-PEDES.}
	\label{tab:DRSDM}
	\begin{tabular}{c|ccc}
		\toprule
		$r$ & Top-1 & Top-5 & Top-10 \\
		\midrule
		$0$ & 57.84 & 79.97 & 87.47 \\
		$0.25$ & 58.61 & \textbf{81.05 }& 87.28 \\
		$0.5$ & \textbf{59.98} & 80.41 & \textbf{87.56} \\
		$0.75$ & 58.48 & 80.18 & 87.29 \\
		$0.9$ & 54.35 & 77.71 & 86.09 \\
		\bottomrule
	\end{tabular}
\end{table}

\begin{table}
	\caption{Performance comparison of Euclidean distance and ranking loss utilized in SDM on CUHK-PEDES.}
	\label{tab:msesdm}
	\begin{tabular}{c|ccc}
		\toprule
		Method & Top-1 & Top-5 & Top-10 \\
		\midrule
		Euclidean Distance & 58.93 & 80.32 & 87.47 \\
		Ranking Loss & \textbf{59.98} & \textbf{80.41} & \textbf{87.56} \\
		\bottomrule
	\end{tabular}
\end{table}

\subsection{Ablation Analysis}
To further investigate the effectiveness and contribution of each proposed component in DSSL, we perform a series of ablation studies on the CUHK-PEDES dataset. The top-$1$, top-$5$ and top-$10$ accuracies (\%) are reported and the best result in each table is presented in bold. As shown in Table \ref{tab:alignment}, comparing with a baseline which is proposed following IMG-Net \cite{wang2020img} without the Inner-Modal Self-Attention Module, DSSL achieves superior performance on both CUHK-PEDES \cite{Shuang2017Person} and our proposed RSTPReid with the aid of proper surroundings-person separation. Additionally, images of each person in RSTPReid are caught by different ones out of 15 independent cameras in both indoor and outdoor scenarios in various periods of time and thereby differ in illumination condition, weather, view angle, body position, etc., which makes RSTPReid obviously a much more challenging benchmark, on which the retrieval performance stumbles, hence leaving much space for further research.

\subsubsection{Surroundings-person separation and fusion mechanism}
\label{sec:exp_SPSM+SPFM}

As shown in Table \ref{tab:sp}, the retrieval result in the first row is given by a model without the surroundings-person separation and fusion mechanism ($SPSM + SPFM$) along with the mutually exclusion constraint ($MEC$). It directly mapping multi-modal data into a latent common space as many of the existing methods do. The top-$1$, top-$5$ and top-$10$ performances drop sharply by $4.46\%$, $2.79\%$ and $2.37\%$, respectively, which demonstrates that our proposed method is more able to properly catch discriminative clues about the corresponding person while drop the misaligned information from complex high-dimensional multi-modal data than unconstrained mapping paradigms. By merely utilizing a $SPSM + SPFM$ mechanism, without a mutually exclusion constraint, the top-$1$, top-$5$ and top-$10$ performances improve by $1.79\%$, $1.71\%$ and $1.67\%$, respectively, which further proves the validity of $SPSM + SPFM$. However, the performance is still $2.67\%$, $1.08\%$ and $0.70\%$ respectively worse than the complete DSSL. This suggests that without the mutually exclusion constraint to ensure the orthogonality between person and surroundings features, information is not able to be well distributed between them. In Table \ref{tab:catadd}, the addition and concatenation methods for combing the two input vectors before handled by the MLP in SPFM are compared. It turns out that the two methods give similar results, with the addition method slightly better.

\begin{figure}[h]
	\centering
	\includegraphics[width=\linewidth]{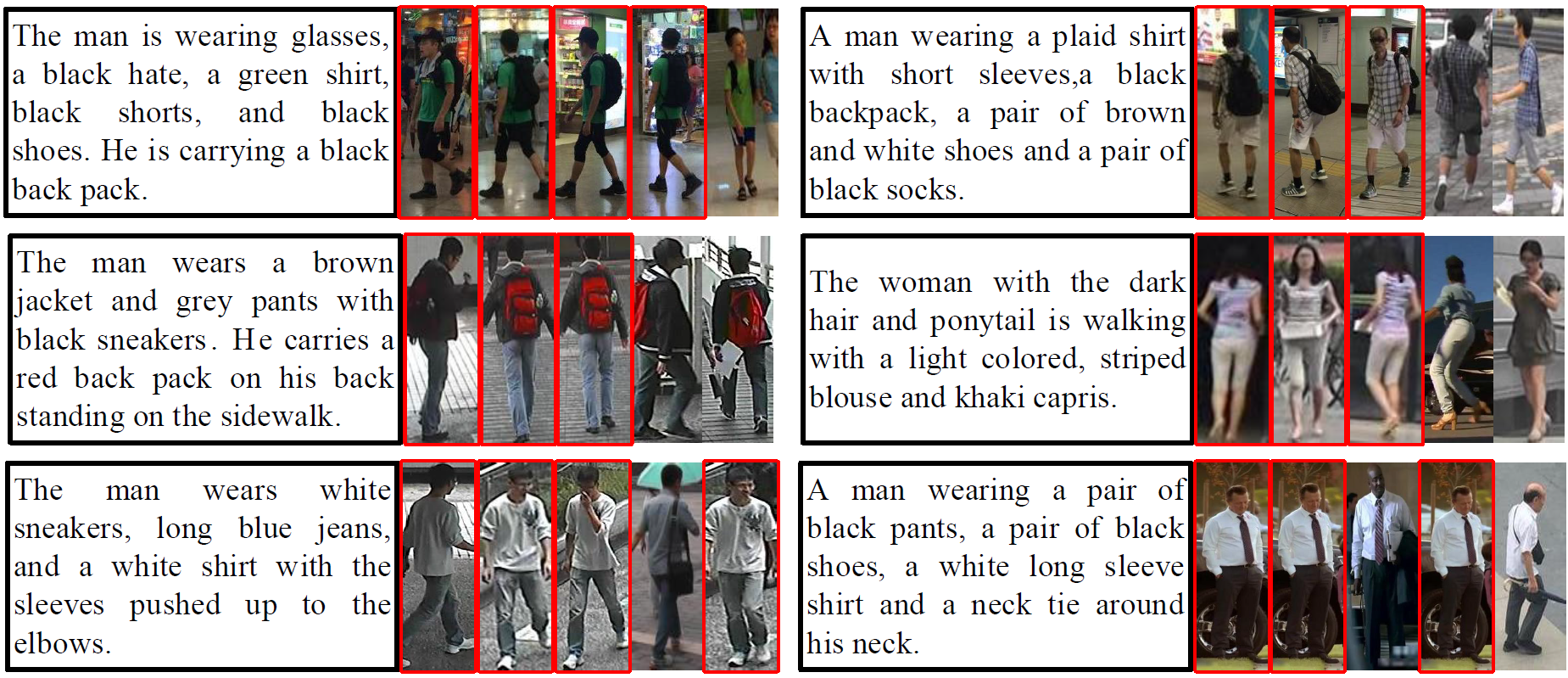}
	\caption{Examples of top-5 text-based person retrieval results by DSSL. Images of the target pedestrian are marked by red rectangles.}
	\Description{Six examples of top-5 text-based person retrieval results by DSSL.}
	\label{fig:results}
\end{figure}

Some examples of the top-$5$ text-based person retrieval results by DSSL are shown in Fig. \ref{fig:results}. Images of the target pedestrian are marked by red rectangles. As can be seen in the figure, many of the pedestrians in mismatched person images also look quite similar to the target one, which is consistent with the distribution pattern of features discussed above. It seems necessary to find ways to dig deeper into the semantic information and draw similar clusters closer without mixing with each other, which remains for our future work.

\subsubsection{Alignment paradigms}
To adequately utilize multi-modal and multi-granular information for a higher retrieval accuracy, five different alignment paradigms are adopted. Extensive ablation experiments are conducted on both CUHK-PEDES and RSTPReid to prove the effectiveness of them and the results are reported in Table \ref{tab:alignment}. The results show that utilizing more than one single alignment brings performance gain, which indicates that the use of multi-modal and multi-granular features in DSSL can provide more comprehensive information, hence leading to a more accurate retrieval. By combining $Align II$ and $Align III$ with $Align I$, SPSM can more completely separate person and surroundings information with the aid of SPFM and PDM. Besides, comparing the third row from the bottom with the last row in Table \ref{tab:alignment}, the top-$1$, top-$5$ and top-$10$ performance increase by $1.12\%$, $0.71\%$, $0.61\%$ and $1.43$, $1.25$, $0.56$ respectively on CUHK-PEDES and RSTPReid after the two fine-grained alignments $Align IV$ and $Align V$ are added, which reals the effect of utilizing multi-granular clues.

\begin{table}
	\caption{Comparison with other state-of-the-art methods on CUHK-PEDES.}
	\label{tab:sota}
	\begin{tabular}{l|ccc}
		\toprule
		Method & Top-1 & Top-5 & Top-10 \\
		\midrule
		CNN-RNN \cite{reed2016learning} & 8.07  & -  & 32.47 \\
		Neural Talk \cite{vinyals2015show} & 13.66  & -  & 41.72 \\
		GNA-RNN \cite{Shuang2017Person} & 19.05  & -  & 53.64 \\
		IATV \cite{li2017identity} & 25.94  & -  & 60.48 \\
		PWM-ATH \cite{Chen2018} & 27.14  & 49.45  & 61.02 \\
		Dual Path \cite{zheng2020dual} & 44.40  & 66.26  & 75.07 \\
		GLA \cite{chen2018improving} & 43.58  & 66.93  & 76.26 \\
		MIA  \cite{niu2020improving} & 53.10 & 75.00 & 82.90 \\
		A-GANet \cite{mm2019graphreid} & 53.14 & 74.03 & 81.95 \\
		GALM \cite{Jing2018Pose} & 54.12 & 75.45 & 82.97 \\
		TIMAM \cite{ARL} & 54.51 & 77.56 & 84.78 \\
		IMG-Net \cite{wang2020img} & 56.48 & 76.89 & 85.01 \\
		CMAAM \cite{aggarwal2020text} & 56.68 &	77.18 &	84.86 \\
		HGAN \cite{zheng2020hierarchical} & 59.00 & 79.49 & 86.6 \\
		NAFS \cite{gao2021contextual} & 59.94 & 79.86 & 86.70 \\
		\textbf{DSSL (ours)} & \textbf{59.98} & \textbf{80.41} & \textbf{87.56} \\
		\midrule
		NAFS + RVN \cite{gao2021contextual} & 61.50 & 81.19 & 87.51 \\
		\textbf{DSSL + RR (ours)} & \textbf{62.33} & \textbf{82.11} & \textbf{88.01} \\
		\bottomrule
	\end{tabular}
\end{table}

\subsubsection{Signal denoising module (SDM)}
\label{sec:exp_SDM}

Comprehensive experimental analysis is as well carried out to study the proposed signal denoising module (SDM). As shown in Table \ref{tab:DRSDM}, ablation experiments are conducted to search for the optimal zeroing rate $r$. It can be observed that initially the performance of DSSL follows a increasing tendency with the growth of $r$. After reaching a peak, the performance begins to turn worse as $r$ continues to go larger. It is conceivable that by randomly dropping a certain number of elements in the input vector at random and then reconstructing it following a autoencoder manner, which is required to be well matched with the original feature under a ranking loss, redundant noise signals are inclined to be removed. Note that when $r$ is set to $0$, there is no zero setting process before the input vector is reconstructed. With the growth of $r$, SDM gradually finds an optimal zeroing rate that the noise signal are just properly dropped while person information is well retained, which gives a summit in performance. After bypassing the summit, an excess of amount of information will be discarded, and hence the retrieval performance will undoubtedly go down. As can be seen in Table \ref{tab:DRSDM}, when $r$ reaches $0.9$, the accuracies of top-$1$, top-$5$ and top-$10$ all fall sharply.

Besides, we compare the performance of utilizing Euclidean distance and ranking loss in SDM (shown in Table \ref{tab:msesdm}). The top-$1$ accuracy of DSSL with ranking loss in SDM is $1.05\%$ higher than the one with Euclidean distance, which indicates that rather than the commonly used Euclidean distance for reconstruction, ranking loss is better at dealing with the particularity of retrieval problems. We also train and evaluate DSSL without the whole SDM (shown in Table \ref{tab:sp}). The top-$1$, top-$5$ and top-$10$ performance drop by $2.52\%$, $0.74\%$ and $0.37\%$ respectively, which reveals the effect of SDM as well.

\subsubsection{Salient attention module (SAM)}
As shown in Table \ref{tab:sp}, the top-$1$ accuracy drops by $2.03\%$ without the salient attention module (SAM) which utilize the extracted person information to highlight and catch body part information in the visual local features. The results indicate the effectiveness of SAM.

\subsection{Comparison With Other State-of-the-art Methods}
Table \ref{tab:sota} shows the comparison of DSSL against $15$ previous state-of-the-art methods including CNN-RNN \cite{reed2016learning}, Neural Talk \cite{vinyals2015show}, GNA-RNN \cite{Shuang2017Person}, IATV \cite{li2017identity}, PWM-ATH \cite{Chen2018}, Dual Path \cite{zheng2020dual}, GLA \cite{chen2018improving}, MIA \cite{niu2020improving}, A-GANet \cite{mm2019graphreid}, GALM \cite{Jing2018Pose}, TIMAM \cite{ARL}, IMG-Net \cite{wang2020img}, CMAAM \cite{aggarwal2020text}, HGAN \cite{zheng2020hierarchical} and NAFS \cite{gao2021contextual} in terms of top-$1$, top-$5$ and top-$10$ accuracies in the text-based person retrieval task. Our proposed DSSL achieves $59.98\%$, $80.41\%$ and $87.56\%$ of top-$1$, top-$5$ and top-$10$ accuracies, respectively. It can be observed that DSSL outperforms existing methods, which proves the effectiveness of our proposed method. Both with a cross-modal re-ranking method, DSSL outperforms NAFS as well. With person and surroundings information separated properly, DSSL surpasses methods which directly map data into a common space. Moreover, compared to methods building similarities based on attention mechanism, DSSL achieves a significant performance improvement, which indicates that our proposed surroundings-person separation mechanism is more able to properly capture detailed person information.

\section{Conclusion}
In this paper, we propose a novel Deep Surroundings-person Separation Learning (DSSL) model to effectively extract and match person information, and hence achieve a superior retrieval accuracy.  A surroundings-person separation and fusion mechanism plays the key role to realize an accurate and effective surroundings-person separation under a mutually exclusion constraint. In order to adequately utilize multi-modal and multi-granular information for a higher retrieval accuracy, Five diverse alignment paradigms are adopted. Extensive experiments are carried out to evaluate the proposed DSSL on CUHK-PEDES, which is currently the only accessible dataset for text-base person retrieval task. DSSL outperforms previous methods and achieves the state-of-the-art performance on CUHK-PEDES. To properly evaluate the proposed method in the real scenarios, a Real Scenarios Text-based Person Reidentification (RSTPReid) dataset is further constructed to benefit future research on text-based person retrieval.


\vfill\eject 
\bibliographystyle{ACM-Reference-Format}
\balance
\bibliography{jonniewayy}

\appendix

%
%
%
%
%
%
%

\end{document}